\documentclass[sigconf]{acmart-me}

\usepackage{booktabs} 
\usepackage{url}
\usepackage{color}
\usepackage{enumitem}
\usepackage{subcaption}
\usepackage{balance}

\setcopyright{rightsretained}

\acmDOI{}

\acmISBN{}

\acmConference[MediaEval'20]{Multimedia Evaluation Workshop}{December 14-15 2020}{Online} 
\acmYear{2020}
\copyrightyear{}

\acmPrice{}

\begin{document}
\title{Floods Detection in Twitter Text and Images}

\author{Naina Said\textsuperscript{1}, Kashif Ahmad \textsuperscript{2}, Asma Gul\textsuperscript{3}, Nasir Ahmad \textsuperscript{1} Ala Al-Fuqaha \textsuperscript{2}}
\affiliation{\textsuperscript{1}CSE, University of Engineering and Technology, Peshawar, Pakistan \textsuperscript{2} Division of Information and Computing Technology, College of Science and Engineering, Hamad Bin Khalifa University, Qatar Foundation, Doha, Qatar. \\ \textsuperscript{3}Department of Statistics, Shaheed Benazir Bhutto Women University, Peshawar, Pakistan}
\email{{kahmad,aalfuqaha}@hbku.edu.qa,asmagul@sbbwu.edu.pk}  \email{{nainasaid,n.ahmad}@uetpeshawar.edu.pk }

%
%
%
%
%

\renewcommand{\shortauthors}{N. Said et al.}
\renewcommand{\shorttitle}{Flood-related Multimedia}

\begin{abstract}
In this paper, we present our methods for the MediaEval 2020 Flood Related Multimedia task, which aims to analyze and combine textual and visual content from social media for the detection of real-world flooding events. The task mainly focuses on identifying floods related tweets relevant to a specific area. We propose several schemes to address the challenge. For text-based flood events detection, we use three different methods, relying on Bog of Words (BOW) and an Italian Version of Bert individually and in combination, achieving an F1-score of 0.77\%, 0.68\%, and 0.70\% on the development set, respectively. For the visual analysis, we rely on features extracted via multiple state-of-the-art deep models pre-trained on ImageNet. The extracted features are then used to train multiple individual classifiers whose scores are then combined in a late fusion manner achieving an F1-score of 0.75\%. For our mandatory multi-modal run, we combine the classification scores obtained with the best textual and visual schemes in a late fusion manner. Overall, better results are obtained with the multimodal scheme achieving an F1-score of 0.80\% on the development set.   
\end{abstract}

%
%
%
%
%


\maketitle

\section{Introduction}
\label{sec:intro}
Social media outlets, such as Facebook, Twitter, and Instagram, allow users to create, obtain and share instant information. Being an instant source of information, social media outlets especially Twitter has been widely exploited for information gathering and dissemination especially in adverse events, where instant access to relevant information is more crucial \cite{ahmad2018social,ahmad2017jord}. The literature reports several situations where the news agencies are unable to provide timely and accurate information, and shows how social media information on early damage caused by adverse events could help under such circumstances \cite{Said2019}.  

Being one of the most frequently occurred natural disasters, flood events detection in social media has been the focus of the research community over the last few years. For instance, the research work presented in \cite{imran2014aidr} assesses the informativeness of a tweet in the event of an earthquake using machine learning techniques. Similarly, in another study \cite{li2015twitter}, the authors analyze domain adaptation classifiers by utilizing the labeled data from a past disaster event and unlabelled data from a current event. Flood events detection has been also part of the MediaEval challenge for the last four years where each time a different aspect of flood events has been targeted. This year, the task is focused on the detection of flood events relevant to a specific area \cite{andreadis2020floodmultimedia}. In the task, the participants are provided with a collection of Tweeter data containing a large number of tweets' text and associated images, and are asked to develop a multi-modal system capable of automatically detecting flood related events that occurred in a particular area in Italy. It is to be noted that the tweets are provided in the Italian language. 

This paper provides a detailed description of the methods proposed by team UEHBKU for the task. In total, we submitted five runs including a late fusion based multimodal, one image-based, and three textual information-based solutions as detailed in the next section.    

\section{Proposed Approach}
\subsection{Image-based Floods Detection}
For image-based floods detection, based on our experience on a similar type of task \cite{ahmad2018comparative,said2018deep}, we rely on multiple pre-trained CNNs to extract the object-level features from the images, which are then used to train multiple SVM classifiers. The SVM classifiers provide the results in terms of posterior probabilities, which are then combined in a late fusion manner by aggregating the scores. A label with the highest aggregate is selected as the final outcome of the framework. In total, we used three different models namely (i) DenseNet \cite{huang2017densely}, (ii) VggNet-19 \cite{simonyan2014very}, and  ResNet \cite{he2016deep}. It is to be noted that all the models are pre-trained on ImageNet \cite{deng2009imagenet}, and expected to extract object-level features. 

Moreover, to deal with the class imbalance problem, we use Synthetic Minority Oversampling Technique (SMOTE) \cite{chawla2002smote} to synthesize new examples of the rare class. During the oversampling process, the rare class has been increased by a factor 3 to have an equal number of samples in both classes.

\subsection{Text-based Floods Detection}
For text-based analysis of the tweets, two different methods, namely (i) BoW, and (ii) state of the art BERT model \cite{devlin2018bert}, are used to obtain feature vectors from the tweets. The BoW model represents text by describing the occurrence of words within a document where each word count is considered as a feature. BERT, on the other hand, applies bidirectional training of Transformer, which is a popular attention model, to language modeling. It is to be noted that different variations of the BERT model are available. Since the tweets provided for the task are in the Italian language so we utilize the Italian version of the model. Before training the models, the text is cleaned by using some pre-processing techniques to remove punctuation keys, such as commas,full-stops, emojis, URLs, and stop words from the tweets. Similar to the image-based solution, we relied on SMOTE to tackle the class imbalance problem in the textual data. 

The feature vector obtained with BoW is used to train a Naive Bayes classifier where as a logistic regression model is trained on the BERT features. The classification scores obtained with the both individual models are then combined in a late fusion manner by aggregating the probabilities obtained with both models for the final decision.  

\subsection{Multi-modal Analysis}
For the mandatory multi-modal run, the visual and textual information are combined in a late fusion scheme by aggregating the probabilities obtained with the individual models trained on visual and textual features. It is to be noted that in the current implementation, all the models are treated equally by assigning them equal weights. In the future, we aim to use more sophisticated fusion methods by assigning merit-based weights to the models. 

\section{Results and Analysis}
We submitted five different runs for the task where the first three runs are based on the text while Run 4 and Run 5 are based on visual and multi-modal information, respectively. In Run 1, we used BoW for text representation to differentiate between flooded and non-flooded events on Twitter. In Run 2, we relied on a multilingual BERT model to obtain a feature vector for the tweets, and a logistic regression model is then trained on the generated word embeddings. 

The variation in the performance of the models motivated us for the joint use of the models in a late fusion manner for our third run. However, lower than the best individual model' performance (i.e., BoW) is observed for the joint use of the models indicating that BERT is not suited well in our case.

Our Run 4, which is based on visual information only, is motivated by our previous experience \cite{ahmad2017convolutional}, where we combined multiple state-of-the-art pre-trained models in a late fusion manner by aggregating the posterior probabilities obtained with the individual models. 

In our final run, we enrich the textual information with the images associated with each tweet for accurate classification of the tweets. Again, a late fusion method by aggregating the posterior probabilities is utilized to combining the complementary information obtained with text and associated images.  

As can be seen in Table \ref{NITD_results}, overall, better results have been obtained with the multimodal approach, which indicates the superiority of the joint use of textual and visual information for the task. In the case of individual models trained on a single type of feature (i.e., textual or visual), better results have been observed for BoW on the textual features. However, comparable results have been observed for the joint use of the different deep models on visual information.  

Moreover, as can be seen in Table \ref{NITD_results}, the average score of all the teams is very low, which shows the complexity of the task. However, the scores on the development set are reasonably good, which indicates the issues with the test set especially because the teams' average scores for most of the runs are below 20\%.  

\begin{table}[!htb]
    \caption{Evaluation of our proposed approaches in terms of F1-scores. \textit{The first and second columns represent our results on the development and test sets, respectively. For comparison against the other teams, we also provide the average score of all the teams on each run in the third column.}}
    \vspace{-10px}
      \centering
\label{NITD_results}
\begin{tabular}{|c|c|c|c|}
\hline
\textbf{Run} & \textbf{Dev. Set}& \textbf{Test Set}& \textbf{Teams' Average Results} \\ \hline
Run 1 & 0.77 & 0.437 & 0.357  \\ \hline
Run 2 & 0.68 & 0.276 &0.156  \\ \hline
Run 3 & 0.70 & 0.343 &0.199  \\ \hline
Run 4 & 0.75 & 0.129 &0.131  \\ \hline
Run 5 & 0.80 & 0.093 &0.221  \\ \hline
\end{tabular}
\vspace{-10px}
\end{table}

\section{Conclusions and Future Work}
The 2020 Mediaeval flood-related multimedia task was concerned with analyzing Twitter data for flood detection. The goal of the task was to combine the textual and visual information from Twitter in order to develop an automatic classification system to indicate whether a particular tweet's text and the associated image is relevant to an actual flooding event or not. We proposed five different solutions including a multimodal, a visual information based solution, and three text-based methods. We observed that both types of data complement each other, and indeed improves the overall accuracy. In the present study, we performed late fusion using equal weights for all the models. However, in the future, we would investigate different optimization techniques, such as Particle Swarm Optimization and Genetic Algorithm, for assigning merit-based weights to the models in fusion. In addition, we will also explore other text-based models specifically for the Italian language to improve the accuracy of Italian text classification. 
\balance

\bibliographystyle{ACM-Reference-Format}
\def\bibfont{\small} 
\bibliography{sigproc} 

\end{document}